\documentclass[preprint,3p]{elsarticle}
\usepackage{amssymb}
\usepackage{amsfonts,amsmath,wasysym}	
\usepackage[nameinlink]{cleveref}
\usepackage{subcaption}

\begin{document}

\begin{frontmatter}



\title{Bayesian Additive Regression Networks}


\author[am,ddl]{Danielle Van Boxel}
\ead{vanboxel@math.arizona.edu}
\ead[url]{https://dvbuntu.github.io/}

\affiliation[am]{organization={Applied Math GIDP, University of Arizona},%Department and Organization
            addressline={617 N. Santa Rita, Room 410}, 
            city={Tucson},
            postcode={85721}, 
            state={AZ},
            country={USA}}

\affiliation[ddl]{organization={Data Diversity Lab, University of Arizona},%Department and Organization
            %addressline={1103 E. 2nd Street, Harvill Suite 320}, 
            city={Tucson},
            %postcode={85721}, 
            state={AZ},
            country={USA}
            }

\begin{abstract}

We apply Bayesian Additive Regression Tree (BART) principles to training an ensemble of small neural networks for regression tasks. Using Markov Chain Monte Carlo, we sample from the posterior distribution of neural networks that have a single hidden layer. To create an ensemble of these, we apply Gibbs’ sampling to update each network against the residual target value (i.e. subtracting the effect of the other networks). We demonstrate the effectiveness of this technique on several benchmark regression problems, comparing it to equivalent shallow neural networks, BART, and ordinary least squares. Our Bayesian Additive Regression Networks (BARN) provide more consistent and often more accurate results.  On test data benchmarks, BARN averaged between 5\% to 20\% lower root mean square error.  This error performance does come at the cost, however, of greater computation time.  BARN sometimes takes on the order of a minute where competing methods take a second or less.  But, BARN \emph{without} cross-validated hyperparameter tuning takes about the same amount of computation time as tuned other methods.  Yet BARN is still typically more accurate.
\end{abstract}




\end{frontmatter}


\section{Introduction}
In recent years, methods like Bayesian Additive Regression Trees (BART) have shown themselves capable of producing low prediction error on benchmark machine learning problems.  BART uses an ensemble of decision trees, where each one is iteratively updated by sampling from a posterior distribution \citep{chipman2010bart}.  BART originally proved superior to modest neural networks in its time.  Today, however, Neural Networks have come to the forefront of machine learning research \citep{schmidhuber2015deep}.  Thus, we consider adapting the MCMC optimization tactics of BART into Bayesian Additive Regression Networks (BARN) in order to model even more accurately.

One powerful generalization of BART adapted it to a much broader set of problem types, not just regression and binary classification \citep{linero2022generalized}.  Standard BART relies on the particular structure of certain loss functions having closed-form posterior integrals in order to avoid the dimension changing from tree to tree.  In a more recent study applying BART generically to problems like classification and count prediction, however, no closed form is required.  The dimension changes as expected, but they account for this with Reversible Jump Markov Chain Monte Carlo (RJMCMC) \citep{green1995reversible}.  With decision trees, it's possible to account for this in a general way (aided by considering only data points that bin into a particular leaf node).  This expands the BART approach to new classes of machine learning problems.

One limitation, however, of using decision trees as the ensembled component in BART is that its output is inherently stepwise.  That is, small changes in the input vector can lead to discontinuous changes in output.  Even with an ensemble of trees and sampling over the posterior, the prediction function is not, in general, continuous.  SoftBART addresses this by treating tree splits as random rather than deterministic \citep{linero2018bayesian}.  They note that this complicates the fitting procedure as SoftBART computes likelihoods over all leaf nodes rather than a single one.  Another alternative to this is Model Trees BART (MOTR-BART) \citep{prado2021bayesian}.  Here, one replaces the constant leaf output values with a small linear function of the inputs.  MOTR-BART still has stepwise output, but it is stepwise linear rather than stepwise constant.  And as this output is linear, the posterior still has a closed form via a Gaussian integral.  MOTR-BART also helps reduce the number of trees.  Each MOTR-BART tree can capture linear behavior, not just constant behavior.  But in either model, the output can be continuous, matching many real problems we would like to model.

While neural networks dominate popular thinking in machine learning, especially with user interfaces like ChatGPT (and its predecessor, InstructGPT) \citep{ouyang2022training}, they have proven themselves in a variety of research contexts as well.  In computer vision, for example, convolutional neural networks can provide not only high accuracy classification, but also a degree of explainability of portions of images that represent a particular class via techniques like Grad-CAM \citep{selvaraju2017gve}.  Researchers also make neural networks work together in ensembles.  An interesting example of dynamic weighting of multiple networks helped address weather forecasting \citep{maqsood2004ensemble}.  Such ensembling continues to see applied use more recently as well \citep{mongan2023ensemble,alam2020dynamic,vallabhajosyula2022transfer}.  From all this, we can see that neural networks are a competitive algorithm in the ecosystem of machine learning approaches.

Decision trees as a backbone for BART are attractive because of the statistical rigor of a fully specified closed-form posterior.  But with the increased demonstration of machine learning capabilities of neural networks \citep{schmidhuber2015deep}, we go beyond MOTR-BART's extension of linear outputs \citep{prado2021bayesian} and replace decision trees entirely with neural networks.  Using neural networks in this way changes the posterior calculation to no longer have a closed form.  But we can still approximate the posterior and empirically evaluate such Bayesian Additive Regression Networks.

To adapt BART this way, we replace the standard posterior analysis with a revised one.  We refer readers to \citeauthor{linero2022generalized} \cite{linero2022generalized} for a detailed understanding of the MCMC fitting process of BART in a generic model-agnostic way.  This provides background for \Cref{sec:method}, wherein we detail how BARN uses a MCMC fitting procedure inspired by BART.  To show this empirically, \Cref{sec:eval} compares BARN against BART and other methods on some benchmark data sets.  From this, we will see that BARN is highly adaptable and often more accurate than other regression methods.

\section{BARN Model}\label{sec:method}

We adapt the BART procedure \citep{chipman2010bart}, replacing the ensemble of decision trees with an ensemble of neural networks.  Each neural network has a single hidden layer.  That is, we train the ensemble such that $\hat{y}_i = \sum_k M_k(x_i)$, where $M_k$ is the $k$th neural network with some number of neurons and one hidden layer.  In principle the model can be arbitrarily more complicated, but we retain this structure to simplify MCMC calculations, reduce computation time, and limit model overfitting.  Due to the similarities with BART, we call our procedure Bayesian Additive Regression Networks (BARN).

At a high level, we propose architectural changes to a single model, $M_k$, at a time by computing the current residual, $R_k = Y - \sum_{j \neq k} M_j(X)$.  This single model change enables us to condition on the fixed models, $M_{j\neq k}$, to effect Gibbs sampling.  Each MCMC step proposes a modification to $M_k$'s architecture, trains this new model on $(R_k, X)$, and computes an acceptance probability, $A$.  Note that because we determine the number of neurons within each network via the MCMC acceptance/rejection, this process implicitly performs a neural architecture search.  Additionally, we fix the \emph{number of models} in the ensemble as $N$.  The MCMC procedure only modifies existing networks in the ensemble.

To do such modifications with detailed balance, we need the three components of the acceptance probability, $A$.  For the transition probability, $T(M',M)$, we have some flexibility.  While a ``tilted'' distribution would provide faster convergence, it also relies on more knowledge of the target distribution.  As we generally lack such knowledge, we adopt a simple transition rule based on the number of neurons $m'$ and $m$ in the hidden layer of models $M'$ and $M$.  \Cref{eq:barn_prop} specifies the probability of shrinking or growing a network by exactly 1 neuron.  In our analysis, we set $p=0.4$ as the growth transition probability.  This helps regularize the networks to be small, helping avoiding overfitting.  Finally, note that we do not allow networks to have $m=0$ neurons; such transitions are always rejected.

\begin{equation}\label{eq:barn_prop}
T(M',M) = \begin{cases}
			p & m' = m + 1 \\
            1-p & m' = max(1,m-1) \\
\end{cases}
\end{equation}

  Note that we do not consider the weight values themselves in this calculation; doing so would require handling dimensionality changes (i.e. with reversible jump MCMC \citep{green1995reversible}).  And it would require some judgment or heuristic a priori about how the weights should change, \emph{independent of the training process}.  Also, with BART, the distribution of terminal node weights directly leads to a distribution on whole ensemble predictions (as the ensemble is a sum of these known terminal node distributions) \citep{chipman2010bart}.  In a neural network, however, computing the distribution of the output is more involved.  Even if there is a distribution on weights, the distribution of input values is also required.  And interaction between neurons is not independent, as they will have been trained with a traditional optimization.  But such optimization is well studied and experimentally proven to be reliable \citep{kingma2014adam} for neural networks, so we maintain such an approach for BARN.  The rest of the procedure, however, mimics the BART MCMC approach.

When transitioning from $M$ to $M'$, we can make use of the already learned weights of $M$.  If $m' = m+1$, then we transfer the old weights to the new model and randomly initialize the new neuron weights.  This retains previously learned information so the new model need not train exactly from scratch.  Similarly, if $m' = m-1$, then we retain the weights of the first $m-1$ neurons.  Note, however, that these serve only as the initial weights for the network.

% some addition info on mutate + train, maybe move this earlier
Once the transition has been proposed and initial weights set, we can train this proposed network on the current training residual, $R_k$, and $X$.  Note that one can use any number of neural network optimization methods; for simplicity we use the BFGS optimizer \citep{fletcher2000practical} with a maximum of 100 iterations (though this forms an additional hyperparameter to the BARN procedure).

Again, we reiterate that the development of a proposed replacement neural network, $M'$, is a two step process.  First we mutate the architecture by adding or subtracting a node, $m' = m \pm 1$.  And only then do we train the resulting architecture, informed by the original model weights, on the target residual, $R_k$.  This produces a new set of weights, $w_k'$, associated with $M'$, and a new set of model predictions, $M'(X,w_k)$.  

What remains is the posterior, comprised of the evidence and the prior.  Recall that in BART, this is an integral over all possible output values for the given leaf.  The BARN posterior breaks up in a similar way.  The model node count prior, $P(M)$, does not depend on the weights, so it comes out of the integral.  What remains within the integral are the error likelihood (i.e. evidence for the model) and the weight priors, seen simplified in \Cref{eq:barn_post2} and expanded with the normal error as well as flat weight assumptions in \Cref{eq:barn_post3}.

\begin{align}
    P(R_k|M) &= \int P(R_k| M_k,w) P(w) dw \label{eq:barn_post2}\\
P(R_k|M) &= \int \frac{exp(\frac{(R_k-M_k(X,w))^2}{-2\sigma^2})}{\sqrt{2\pi \sigma^2}} 1 dw \label{eq:barn_post3}
\end{align}

Now we come to an issue.  In BART or MOTR-BART, the analogous component is constant, or at least linear, in both cases lending itself to a closed-form integral \citep{chipman2010bart,prado2021bayesian}.  But $M_i(X,w)$ is a complicated nonlinear function of various $w_k$ weights, so we lack such a form.  If we do away with the integral and make the weight sampling part of the MCMC algorithm, then the dimension of each model varies, requiring RJMCMC.  Generalized BART handles something similar with reversible jumps, but they take advantage of the decision tree structure to again find a general closed form even under dimension change.  The weights of a neural network, however, come from an involved training process, in our case BFGS.  This makes obtaining a Jacobian for RJMCMC infeasible.  We cannot compute a BARN posterior this way.

Instead, consider an approximation to the integral over weights.  In neural networks, not unlike decision trees, certain weights produce much higher probabilities.  That is, most of the probability mass for a given weight concentrates around a particular value.  Additionally, swapping neuron positions (e.g. $w_{2i} \to w_{2i+1}$ and vice versa) produces numerically equivalent neural networks.  These equivalent NNs must therefore have multiple strong peaks in the multivariable probability density curves.  Therefore, we approximate the integral itself with the likelihood of the peak (i.e. the integrand at the given weight values).  This does introduce a potential scaling issue, but because we take a ratio of likelihoods (for the old network and the proposed network), these scales should cancel out, leaving us the approximate probability ratio.  So we replace the integral of \Cref{eq:barn_post2} with the approximate \Cref{eq:barn_post_app}.

\begin{equation}\label{eq:barn_post_app}
\int P(R_k|X,M_k) dw \approx \prod_{j \in \text{valid}} \frac{e^{-\frac{1}{2}\left(\frac{(R_k-M_k(X_j))}{\sigma}\right)^2}}{\sigma \sqrt{2\pi}}
\end{equation}

Note that to compute this error likelihood, we apply the models to held-out data (but not the test data), else models may quickly overfit (an open question is if a prior on the error itself can better control overfitting without a held-out data set).  This component then goes into the calculation of the acceptance ratio, $A$.

Next, we still need to specify a prior, $P(M)$, based on the neural network architecture as well as its learned weights.  But as the weights can vary considerably depending on data, we restrict our prior to the size of the network.  While different domains may propose different priors, we again want to encourage relatively small models.  To that end, we encode a model's prior with a Poisson distribution on the count of neurons in its hidden layer.  We choose $\lambda=1$ as the mean of a Poisson distribution after empirically testing a few options (note that we explore more detailed cross-validation for hyperparameter selection in \Cref{subsec:eval_setup}).  So a model, $M$, with $m$ neurons has prior probability:

\begin{equation}\label{eq:barn_prior}
P(M) = \frac{\lambda^m e^{-\lambda}}{m!}
\end{equation}

Finally, we combine the transition, evidence, and prior components into the acceptance ratio calculation.  \Cref{eq:barn_a} shows this for proposing a transition to $M_k'$ from $M_k$ with a residual of $R_k$.

\begin{equation}\label{eq:barn_a}
A(M_k,M_k') = min\left(1, \frac{T(M_k|M_k') P(R_k|X,M_k')P(M_k')}{T(M_k'|M_k) P(R_k|X,M_k)P(M_k)}\right)
\end{equation}

Recall that the residual in \Cref{eq:barn_post_app} are sufficient to replace the contribution from all of the frozen models in the ensemble.  When we fix those models and subtract their effect to obtain the residual, we are implementing a form of Gibbs sampling.  That is, we approximate $P(Y|X,M_1,M_2,\cdots)$ as in \citep{chipman2010bart} (i.e. the entire distribution given all the models, not just the particular model).  Additionally, we sample the model error, $\sigma$, exactly as is done in BART, using an inverse gamma distribution.  By iterating through the models in this fashion, we develop better and better approximations to the posterior.  After some burn-in period, the distribution of models reflects the posterior given the available architectures and the data.  We can then sample one (as for point inference) or more sequences of MCMC ensembles (as for intervals) as the model output.

\section{Method Evaluation}\label{sec:eval}

We first assess the performance of BARN on benchmark regression problems that have been previously used in studies examining the performance of newly-developed methods. Then, we examine how BARN and competing methods perform on a series of controlled synthetic data sets.  This allows us to see how BARN's performance changes in response to specific parameters like increasing noise or number of irrelevant features.

\subsection{Model Evaluation Setup}\label{subsec:eval_setup}

To test the effectiveness of BARN, we train ensembles against nine benchmark regression problems and compare its error to that of existing regression techniques.  Seven of the data sets analyzed in this study are freely available from the UCI Machine Learning repository \citep{Dua:2019}, one is from a recent study \citep{roman2022bayclump}, and the last is a randomly generated linear regression problem with eight relevant and two irrelevant variables.  We compare BARN against BART which previously performed very well on some of these data sets \citep{biau2019neural}.  \Cref{tab:datasets} shows the size of each data set in features and data points.

\begin{table}[htb]
\centering
\caption{Representative data sets for testing BARN}
\begin{tabular}{lrr}
Dataset & Features & Data Points\\ \hline
boston & 13 & 506\\
concrete & 9 & 1030\\
crimes & 101 & 1994\\
diabetes & 10 & 442 \\
fires & 10 & 517\\
isotope & 1 & 700 \\
mpg & 7 & 398\\
random & 10 & 1000 \\
wisconsin & 32 & 194\\
\end{tabular}
    \label{tab:datasets}
\end{table}

While these data sets provide a diverse set of problems on which to test BARN, we also implement a variety of synthetic data sets for additional testing.  We generate this data by starting with different functional relationships, such as a randomly initialized linear relationship $y_i = \beta x_i$.  Another function is pairs of indices of discrete clusters in the input space.  We also extend the clusters by training a random forest on the indices and using its prediction as output.  Finally, we have three multivariate functions from \citet{friedman1991multivariate}.  $F1$ is a 5-dimensional combined polynomial/trigonometric function used to evaluate BART \citep{chipman2010bart}.  $F2$ is a function involving products and reciprocals.  Finally, $F3$ is similar to $F2$, but it is put through an arctangent function.  We summarize these in \Cref{eq:f1} - \Cref{eq:f3}.  This diversity of relationships provides linear, nonlinear, and discontinuous data sets for modeling.

\begin{align}
    F1(X) &= 10 \sin(\pi X_1X_2) + 20(X_3-1/2)^2 + 10X_4 + 5X_5\label{eq:f1}\\
    F2(X) &= \sqrt{X_1^2 + (X_2 X_3  - 1 / (X_2 X_4))^2}\label{eq:f2}\\
    F3(X) &= arctan\left(\frac{X_2 X_3  - 1 / (X_2 X_4)}{X_1}\right)\label{eq:f3}\\
\end{align}

To further diversify our synthetic data, we perturb the output by adding varying amounts of noise (measured by signal-to-noise ratio, or SNR), adding extra irrelevant features to the $X$ vector, or changing the number of overall features.  This creates problems of different complexity and challenge on which to test our methods.  For more detail on this synthetic testing, \Cref{tab:synth_plan} is a summary.

\begin{table}[htb]
\centering
    \caption{Synthetic Data Generation Parameters}
\begin{tabular}{lrr}
Factors & Options \\ \hline
    Functional Relationship & Cluster, Forest, Friedman (1-3), Linear \\
    SNR & 10, 1 \\
    \# Relevant Features & 100, 10 \\
    \% Irrelevant Features Added & 90\%, 10\% \\
\end{tabular}
    \label{tab:synth_plan}
\end{table}

For each data set, before training with our various algorithms, we do minimal preprocessing.  For better comparison across data sets, we rescale outputs to be have mean 0 and variance 1.  For input variables, we perform a principal component analysis (PCA) with the number of dimensions equal to the number of features (i.e. decorrelating but retaining all information).  As we know neural networks benefit from this input transformation \citep{orr1998neural}, we allow this step for all models as would be done in practical applications.

Now we specify the parameters needed for BARN.  We set the number of networks in the ensemble to a small number, 10, even though BART recommends 100-200 trees \citep{chipman2010bart}.  Using only 10 networks instead of 100 helps reduce computation time since we have to train each network at every step.  We also briefly tested using BARN with a single neural network as a form of architecture search.  This generally resulted in slightly worse models (about 2\% worse RMSE than ensembled BARN of equivalent size), so we retain the ensemble approach.  And as noted in \Cref{sec:method}, our growth transition probability is $p=0.4$, and the prior on model size is a Poisson distribution with $\lambda=1$ neuron.  For training the NNs, we use BFGS optimization with up to 100 iterations and include a small $L_2$ regularization penalty of 0.001 on the weight magnitudes.  To initialize the networks, we set the initial neuron count to 1 for all of them (mimicking BART's starting point of single node trees \citep{chipman2010bart}), and train them each independently on $Y/10$ (i.e. assume they all contribute equally to the result).  We run BARN for potentially 200 iterations, always doing at least 20.  This is similar to the Keras \texttt{EarlyStopping} callback \citep{chollet2015keras}.  Every five steps, we check the validation error for improvement.  If it gets worse by more than the tolerance, 0.0001, we stop training early per standard machine learning practice \citep{genccay2001pricing}.  This helps ensure that models have converged.  We accept the ensemble at the end of this procedure as the final one for testing.

\begin{figure}[ht]
\centering
    \includegraphics[scale=0.66]{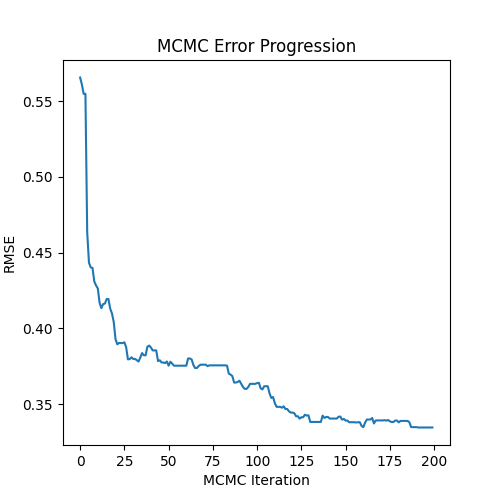}
    \caption{Typical error results for one of the data sets during the MCMC process shows burn-in achieved within the full run}
    \label{fig:phi}
\end{figure}

To confirm that these MCMC parameters are acceptable, we apply a simple batch means analysis test.  In our case, we measure $\phi_t = \sqrt{\sum_i \frac{(y_i-M^t(x_i))^2}{n}}$, the RMSE of the ensemble at each iteration, $t$.  As we use the validation data for the acceptance testing, we also use it for this residual.  Even with our modest amount of training, these values tended to be less than 1\% the size of the RMSE variance across ensembles in independent runs, so we have some confidence that this level of burn-in is sufficient.  Additionally, \Cref{fig:phi} shows a typical error progression indicating convergence in RMSE.  This type of convergence is often checked for ``early stopping'' of an algorithm \citep{prechelt2002early}.  Note that the RMSE converging means the error of the ensembles has converged, not that the parameters such as weights have converged within the MCMC.  Because the networks transition, we don't necessarily expect the weights themselves to converge to a value.  Finally, we are most interested in error estimates across data sets, so we limit discussing details about confirming convergence to just the overall RMSE approach.

For comparison to BARN, we train a single large neural network, a large BART forest, and an OLS model.  For the large neural network, we retain a single hidden layer, but we set the number of neurons equal to the total neuron count in the BARN ensemble.  Such a network has the same number of weights ((number of inputs+1) $\times$ number of neurons).   We train this network for 2000 epochs using the same optimization and regularization parameters as used in BARN.  For BART, we run with default parameters with \texttt{wbart} in the BART package (100 trees, $k=2, \nu=3, q=0.99$)  as those have been shown to perform well \citep{chipman2010bart}.  This already encourages small trees, and BART is fast enough that the default burn-in of 1000 iterations is sufficient and computable.  This provides three alternatives against which to evaluate BARN.

For each method (excluding OLS), we also test a version of it with five-fold cross-validated hyperparameter tuning (denoted with ``CV'' in tables and figures).  For BARN, we explore choices for the prior mean of the model size, $\lambda$; the number of networks in the ensemble; and the activation function in the neural network.  With two options for each, we have a total of eight configurations to test on each of five folds.  For BART, we fix the number of trees at 100, but we vary $k$, $\nu$, and $q$ for a total of 12 combinations.  With the single large neural network, we consider model sizes in different multiples of the BARN-derived size, different learning rates, number of epochs, and like BARN, two options for the activation function.  Table \ref{tab:cv} summarizes the options for each method.  In all cases, we select the set of parameters with the smallest root mean square error.

\begin{table}[htb]
\centering
    \caption{Hyperparameter options for cross-validated methods}
\begin{tabular}{llc}
    Method   &     Parameter & Values \\ \hline \hline
    BARN     &  $n$, Number of Networks & 10, 20 \\
             &  $\lambda$, Poisson Mean Number of Neurons Prior & 1, 2 \\
             &  NN Activation Function & Sigmoid, ReLU \\ \hline
    BART     &  $k$, standard deviation prior on $\sigma_\mu$ & 1,2,3,4,5 \\
             &  $(\nu, q)$, priors on $\sigma$ & ((3,0.9), (3,0.99), (10,0.75))\\ \hline
    Big NN   &  BARN neuron count multiplier & 1, 2, 10 \\
             &  $lr$, Adam initial learning rate & 1e-5, 1e-4 \\
             &  Number of epochs & 2000, 4000 \\
             &  NN Activation Function & Sigmoid, ReLU \\ \hline
\end{tabular}
    \label{tab:cv}
\end{table}

For each data set, we perform 40 random trials with each type of model.  In each trial, we randomly split the data into 50\% training, 25\% validation (for early stopping checking), and 25\% testing, using the same split across model types for consistency.  After completion, we compute the average RMSE of the test predictions.

\subsection{Benchmark Results}\label{subsec:res}

\begin{figure}[ht]
	\centering
	\includegraphics[scale=0.6]{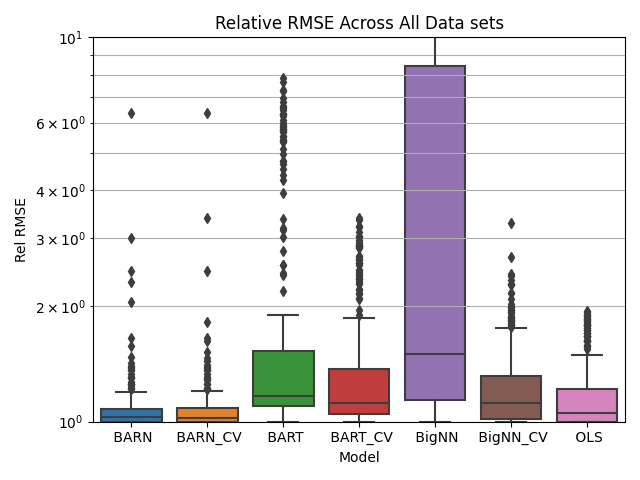}
	\caption{BARN is more adaptable to different problems than other methods and outperforms them across data sets}
	\label{fig:box_results}
\end{figure}

\begin{figure}[ht]
\centering
    \includegraphics[scale=.8]{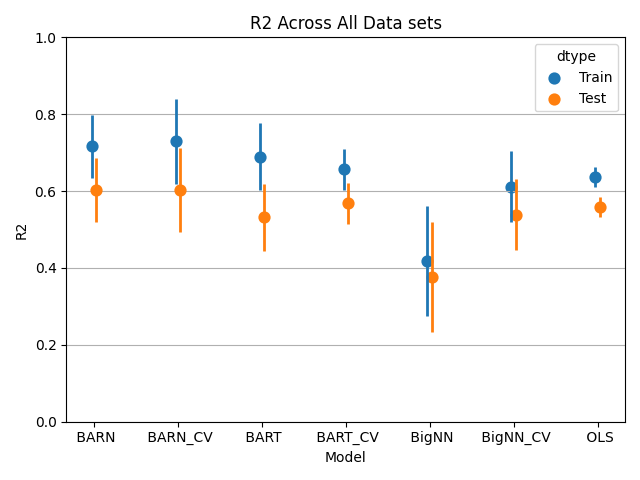}
    \caption{BARN resists overfitting similar to other methods when comparing $R^2$ values across training and testing data with pooled variance}
    \label{fig:r2_results}
\end{figure}

\Cref{tab:results} shows the mean RMSE for each data set and model. We summarize this across data sets into the relative RMSE in \Cref{fig:box_results}, and we compile the median points for both absolute and relative RMSE into \Cref{tab:med_results}, along with pooled sample standard deviations.  Each point in \Cref{fig:box_results} is the test RMSE of that method divided by the test RMSE of the best method for that particular data split.  Additionally, \Cref{tab:max_results} shows the maximum relative error seen for each model and data set.  Note that for BARN, the forest fires data set is the only one with relative error large enough to contain the more severe outliers in \Cref{fig:box_results}.  Finally, \Cref{fig:r2_results} display $R^2$ values of methods averaged across data sets, with error bars showing the pooled variance estimate.  Because different data sets have different amounts of noise, we expect them to have different \emph{mean} $R^2$ values.  By using pooled variance, that is, the variance about \emph{each} mean, we can estimate the variance expected in a typical data set.  Ignoring this and using sample variance would artificially inflate the error bars.  From these metrics, we see that BARN appears to have generally lower relative error and better fit across data sets.

\begin{table}[htb]
\centering
\caption{Mean scaled RMSE over 40 trial with smallest mean error in bold for each data set}
\begin{tabular}{lrrrrrrr}
    Data set   &     BARN & BARN CV & BART & BART CV & Big NN & Big NN CV & OLS\\ \hline

california       & 0.513    &   \textbf{0.495}   & 0.524   &   0.498  &  0.857&    0.566  & 0.558\\
concrete         & 0.375  &   \textbf{0.367} & 0.533&   0.462&  6.266&    0.702   & 0.631 \\
crimes           & 0.622    &   0.627    & 0.708   &   0.649    &  0.736    &    \textbf{0.599}   & 0.614    \\
diabetes         & 0.734  &   0.745 & 0.807  &   0.791 &  0.721  &    0.724 & \textbf{0.719}  \\
fires            & 1.494&   1.441& 1.358  &   1.221&  4.088&    1.130  & \textbf{1.113}  \\
isotope          & 0.299  &   \textbf{0.298} & 0.327 &   0.315 &  0.322 &    0.329 & 0.299 \\
mpg              & \textbf{0.354}  &   0.361& 0.480   &   0.454   & 19.539   &    0.461& 0.408\\
random           & 0.073&   0.074  & 0.386&   0.179  &  0.096&    0.089   & \textbf{0.068}  \\
wisconsin        & 1.123&   1.097& 1.104&   1.058  &  9.028&    \textbf{1.024}    & 1.189\\
\end{tabular}
    \label{tab:results}
\end{table}

\begin{table}[htb]
\centering
\caption{Median test RMSE, relative RMSE, and pooled standard deviation across types of data sets, excluding forest fires data}
\begin{tabular}{lrrrrrrr}
Data set, metric       &     BARN & BARN CV & BART & BART CV & Big NN & Big NN CV & OLS\\ \hline
Real, Median        		 & 0.390 & \textbf{0.383} & 0.517 & 0.477 & 0.732 & 0.570 & 0.562 \\
Real, pooled $\sigma$        & 0.071 & 0.057 & 0.056 & 0.056 & 9.634 & 0.070 & 0.059 \\
Real, rel, Median            & 1.016 & \textbf{1.014} & 1.168 & 1.117 & 1.409 & 1.170 & 1.089 \\
Real, rel, pooled $\sigma$   & 0.096 & 0.062 & 0.391 & 0.156 & 25.789 & 0.173 & 0.079 \\
Synth                		 & 0.788 & \textbf{0.761} & 0.881 & 0.847 & 0.873 & 0.791 & 0.780 \\
Synth, pooled $\sigma$       & 0.083 & 0.082 & 0.068 & 0.073 & 0.078 & 0.118 & 0.059 \\
Synth, rel                   & 1.050 & \textbf{1.026} & 1.206 & 1.158 & 1.207 & 1.053 & 1.060 \\
Synth, rel, pooled $\sigma$  & 0.098 & 0.089 & 0.107 & 0.105 & 0.115 & 0.253 & 0.084 \\
\end{tabular}
    \label{tab:med_results}
\end{table}

\begin{table}[htb]
\centering
\caption{Maximum Relative RMSE for each modeling method and data set}
\begin{tabular}{lrrrrrrr}
    Data set   &     BARN & BARN CV & BART & BART CV & Big NN & Big NN CV & OLS \\ \hline

california		 &    2.31 & 1.09  & 1.21 & 1.11 &  10.96 & 1.34 & 1.32 \\
concrete         &    1.17 & 1.13  & 1.72 & 1.55 &  76.37 & 3.28 & 1.94 \\
crimes           &    1.04 & 1.06  & 1.23 & 1.09 &   1.40 & 1.00 & 1.04 \\
diabetes         &    1.19 & 1.44  & 1.29 & 1.25 &   1.10 & 1.15 & 1.07 \\
fires            &    6.34 & 6.34  & 3.36 & 2.09 &  25.94 & 1.99 & 1.21 \\
isotope          &    1.00 & 1.00  & 1.15 & 1.08 &   1.28 & 1.82 & 1.00 \\
mpg              &    1.10 & 1.40  & 1.67 & 1.57 & 354.35 & 1.76 & 1.37 \\
random           &    1.24 & 1.36  & 7.82 & 3.40 &   1.98 & 1.86 & 1.01 \\
wisconsin        &    1.39 & 1.39  & 1.36 & 1.36 &  48.79 & 1.26 & 1.56 \\
\end{tabular}
    \label{tab:max_results}
\end{table}

\begin{figure*}[ht]
\centering
	\begin{subfigure}[t]{\textwidth}
		\centering
        \makebox[\textwidth][c]{\includegraphics[scale=0.6]{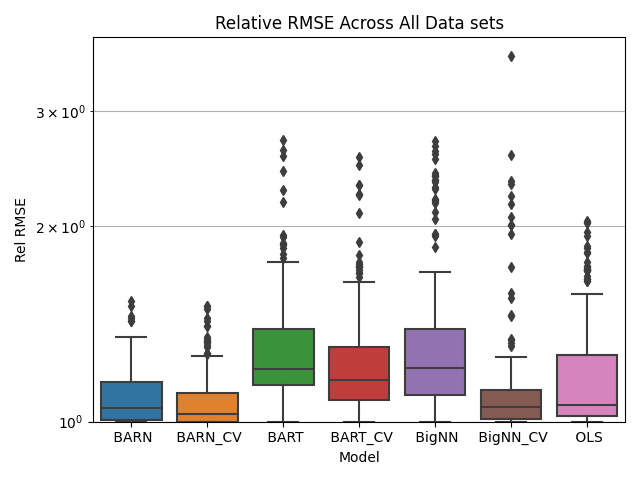}}
        \caption{BARN is adaptable, as seen on a variety of synthetic data sets}
        \label{fig:synth_box_results}
	\end{subfigure}%

	\begin{subfigure}[t]{\textwidth}
		\centering
        \makebox[\textwidth][c]{\includegraphics[scale=.6]{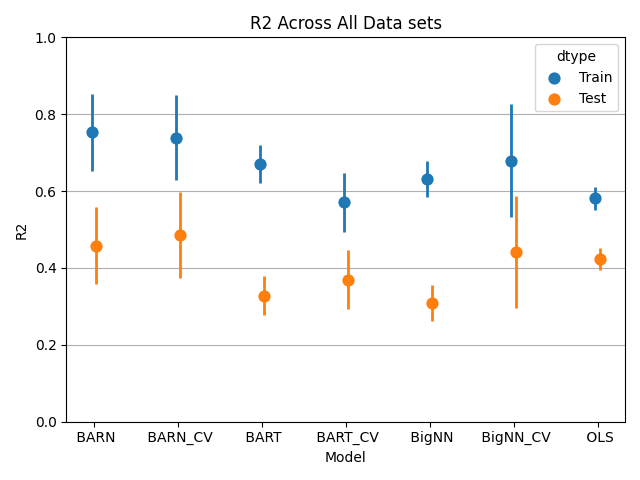}}
        \caption{$R^2$ for synthetic training and testing data shows some overfit}
        \label{fig:synth_r2_results}
	\end{subfigure}
    \caption{BARN is competitive with other methods across data sets}
    \label{fig:synth_data_results}
\end{figure*}

For the synthetic data sets, we perform a similar analysis.  Here, we focus on how changing each parameter affects the results.  We do, however, include the overall relative error and $R^2$ in \Cref{fig:synth_box_results}, where BARN again outcompetes alternatives.  For more detail, we have a set of subplots for each of the variants in \Cref{fig:synth_all_matrix_results} in \Cref{sec:more_synth_reg}.  Across all of these, we can further inspect BARN's robustness.

By examining a broad cross section of real and synthetic data, we can compare BARN to other methods in highly variable situations.  Some of these, like the forest fires data set, have non-normal error terms.  Others, like the Wisconsin breast cancer set, have a very limited amount of data available for training and testing.  What's more, we see from the synthetic data how BARN responds to changing the amount of noise and presence of many features (irrelevant or not).  As visible in the figures, BARN appears competitive with other methods, often outperforming or matching the next-best method.

\subsection{Discussion}\label{sec:dis}

Based on these visualizations, we see that BARN tends to have lower testing error than shallow neural networks, BART, and OLS across data sets.  We choose to recompute BART results rather than using those in \citet{biau2019neural} in order to maintain data set and data split consistency.  The results in \Cref{tab:results} show BARN is often the best performing, with either BARN or tuned BARN (``BARN\_CV'') having the lowest RMSE in four of nine data sets.  In two of these, california and isotope, however, other methods like tuned BART or OLS were within 1\% of BARN's error.  Note that it is possible for the non-cross-validation version to have lower testing error, as the model selection is done on training data.  The relative RMSE shown in \Cref{fig:box_results} displays even more clearly how BARN adapts well across a variety of problems, with BARN having tighter and lower errors than other methods.  \Cref{tab:med_results} makes this quantitative; on real data we see that BARN (even without tuning) has a median testing error about 13\% lower than the next-best method, \emph{which is tuned}.  While BART or OLS may dominate a particular problem, BARN has the smallest relative error across data and a much tighter interquartile range.  It does have a seemingly large number of outliers (similar to BART without tuning), but \Cref{tab:max_results} indicates that these large errors come from the forest fires data set, as the maximum relative error for BARN on other data sets is smaller than the outliers.  In the forest fire data set, the target value is the area of forest burned in a given time period.  Many periods have zero, while others vary, making it zero-inflated or zero-censored data.  Other methods also face challenges with this data, but not always as severely.  Finally, \Cref{fig:r2_results} again shows how BARN finds better fits (in both training and test data) than other approaches.  BARN, tuned or not, has about mean 0.6 $R^2$, while the next highest is tuned BART at 0.58.  These $R^2$ values, however, are within the pooled standard deviation, close enough to not be significantly different.  So although BARN is not universally superior, there is a broad spectrum of situations in which it noticeably outperforms existing methods and may be more generally robust.

We also call out some potential peculiarities in this analysis.  First, the large single NN, without cross-validated tuning, performed very poorly on the real data, with median relative testing error over 50\% higher than BARN.  The available default parameters often failed to converge on various problems, leading to extremely poor results (e.g. ``going off the edge of the plot'' in \Cref{fig:box_results}).  While one could devote more effort to finding better defaults, from the cross-validated tuning, we see that it can perform reasonably well.  And generally, one set of parameters may not transfer well between problem domains.  For BARN, however, cross-validated tuning provides very little advantage.  The view in \Cref{fig:r2_results} perhaps makes this most clear.  There is a small boost in \emph{training} $R^2$ with BARN CV (about 0.07 higher than plain BARN), but this does not lead to much of a boost in \emph{testing} $R^2$ (less than 0.01 improvement).  This increase is perhaps just enough to edge out the default parameters as seen in \Cref{tab:results}, but that tuning may not worth the effort.  Both BARN CV and regular BARN are have $R^2$ values well within a pooled standard deviation of each other.  Cross-validation on BARN set the mean model size prior to be $\lambda=1$, which was the default choice, about 85\% of the time.  Combining these observations about BARN and the large NN provides an additional insight.  A BARN ensemble of shallow neural networks can be trivially rewritten as a single neural network that takes a weighted sum of all the neurons in all the networks.  That is, they are structurally identical.  Yet, BARN's training procedure provides it with robustness comparable to that of cross-validation for the the large NN.  Large NNs have the opportunity to numerically duplicate BARN's results, but BARN is generally superior.

Using the synthetic data sets, we see similar behavior in BARN.  As \Cref{fig:synth_box_results} shows, BARN again has a small relative test RMSE across data sets.  We observe two additional points here.  First, cross-validated hyperparameter tuning does have a modest impact on BARN (about 0.01 improvement on testing data), though perhaps still less so than on other methods.  Even without, BARN is competitive with other methods that use such tuning.  Its mean relative test RMSE was 1.1, and with tuning only 1.08.  The next lowest mean relative test RMSE belonged to the tuned NN at 1.15, about 5\% worse than either BARN method.  Turning to \Cref{fig:synth_r2_results}, we can inspect how BARN handles overfitting on our synthetic data sets.  It has a similar amount of overfitting compared with other nonlinear methods, with an $R^2$ about 0.3 lower on testing data than training.  Not surprisingly, OLS overfits less, with a decrease in $R^2$ of about 0.2.  But BARN, with our without tuning, has the highest mean $R^2$, though error bars overlap with tuned neural networks and OLS.  Overall, BARN still performs well on these synthetic sets.

Examining performance on specific sets, we find a few interesting and distinct results.  For the cluster and forest data sets (which mimic an underlying discontinuous relationship), BARN performs better than other methods when there are more features and noise.  Tuned BARN has a mean RMSE of about 0.8 in these situations, whereas the competing methods are 0.9 or higher.  The testing $R^2$ reflects this as well, where tuned BARN is about 0.35 (still poor, but higher than others).  When $SNR=10$ (i.e. low noise) for these data sets, however, other approaches like neural networks have lower error and higher $R^2$, indicating that perhaps BARN has robustness in this situation, but may not be the most accurate when the problem is easier.  The Friedman data sets have somewhat different perspective.  For $F2$ and $F3$, BARN seems better overall, regardless of noise, feature count, or irrelevant features.  With $F1$, it is only superior for $SNR=10$ cases, especially when there are fewer irrelevant features.  Note that this is the opposite of what we see with the cluster and forest data.  We speculate that the periodic nonlinearity of $F1$ is difficult for BARN to track under noisy conditions.  Finally, we note that in the linear data sets, OLS generally performs best, as expected.  More detailed subplots demonstrating these finds are available in \Cref{fig:synth_all_matrix_results}.  Having these synthetic data sets where we know the underlying parameters enables to see that BARN is robust under a variety of conditions where other approaches may have difficulty.

From all these plots and analyses, we see that BARN is effective compared to other methods on many different domains.  We briefly note here, however, that this performance does take additional computation time.  While we save detailed discussion of timing results for another study, BARN without cross-validated hyperparameter tuning takes comparable time to BART or neural networks \emph{with} such tuning.  Yet, even the untuned BARN shows low testing errors in many situations.  It brings together the strengths of both BART-style MCMC model search and modern neural network implementation to produce a better machine learning method.

\section{Summary}\label{sec:fut}

Though \Cref{sec:eval} shows BARN to be competitive, there remains a significant amount of future work on better theoretical understanding and practical implementation.  Most of the theoretical advances relate to more rigorous analysis of the MCMC components.  For example, our selection of transition probabilities was somewhat arbitrary.  While this limitation was also in Bayesian CART \citep{chipman1998bayesian} regarding tree growth and pruning, a follow-up study on generalized BART derived rigorous normal distributions for transition probabilities \citep{linero2022generalized}.  We may be able to derive a similar distribution to improve convergence and reduce variance in the produced models.  Similarly, we need better justification for our model size prior.  While a Poisson or negative binomial distribution is probably sufficient, a heuristic to guide both the distribution and parameters would be helpful.  From cross-validation, we found a mean of $\lambda=1$ for our Poisson prior was most common, so we recommend this as a starting point.  Next, we assumed the errors for each model are independently identically distributed normal, but this is problem dependent.  One can account for relaxations to this by modifying the $P(Y|X,M)$ expression, but it is not clear if we can derive an approximation for this similar to BART's closed forms.  And finally, we should seek either an RJMCMC-compatible form of model proposal with changing dimensions or a better justification for avoiding this.  The approximation of \Cref{eq:barn_post_app} was a start, but one may be able to better account for changing likelihood functions.  Alternatively, a fixed-dimension MCMC may become feasible by setting a maximum network size and keeping most of the weights zero (effectively deleting most of the neurons).  Or, one might propose weights entirely via the MCMC process, completely eschewing traditional weight optimization approaches entirely (as BART does).  We believe our theoretical approach is sound and confirmed by the experimental results, but this remains an open area of research.

%TODO talk about classification briefly, reference that chapter
%Broadening our view further, we can also extend BARN to classification tasks.  Researchers have recently expanded BART to multinomial logistic (rather than probit) outcomes \citep{xu2021inference}; we should be able to apply that technique to BARN if not adapt the additive regressors directly to the various logit estimations (essentially aggregating the residuals into the logit bias terms).  This would open a new class of problems to modeling by BARN.

On the applied side, there is also fruitful future research.  Still somewhat theoretical, but with direct applications, is a version of BARN suitable for classification problems.  BART provides an approach to this using binary probit \citep{chipman2010bart}, which we can adapt to BARN.  Next, making BARN more accessible via simplified software interfaces and additional languages will help it reach more researchers and data scientists.  This also means adapting BARN to a wider variety of backbones.  To illustrate: BART uses decision trees and BARN uses neural networks, but a generic Bayesian Additive Regression \emph{Model} could use, say, support vector machines (SVMs) \citep{hastie2009elements}.  SVMs, in particular, are attractive as they may have fewer dimension-changing issues if a single type of kernel is considered.  An alternative different application is to focus more on the architecture search rather than the ensembling.  If we restrict the number of networks to exactly one, then BARN transition proposals and MCMC acceptance sample from the posterior on the total number of neurons.  With a single hidden layer in the neural net, such an approach may be less effective than exhaustively trying many different neuron counts.  But if we expand the neural network architecture to allow proposing additional layers, or even different types of layers, such as convolutional ones in an image processing context, a BARN-guided search may become more competitive than exhaustion or even genetic algorithm methods \citep{idrissi2016genetic}.  Such extensions to BARN may make it more applicable on ever more diverse problems.

We have shown BARN to accurately model a variety of regression problem domains.  By adapting the MCMC model sampling process of BART to modern neural networks, we employ the strengths of both.  Though our posterior is an approximation, empirical benchmarking shows how BARN can achieve lower test errors on real data sets.  We further demonstrate this effectiveness on synthetic data, setting various parameters like noise and number of features, which are not controllable in real data.  BARN may take additional processing time, but it may not require much hyperparameter tuning, making its overall computation time comparable to other methods like BART and pure neural networks.  Therefore, BARN is a machine learning approach we should consider applying to new data science problems when seeking lower error rates.

\section*{Code Availability}

Software for running BARN in general is available under \texttt{barmpy} on PyPi \cite{pypi}.  Code specific to our analysis is available on request to the author.

\section*{Acknowledgements}

Danielle Van Boxel was supported as a Graduate Research Assistant advised by Dr. Cristian Rom\'an-Palacios in the Data Diversity Lab within the School of Information at the University of Arizona.  Dr. Xueying Tang from the Department of Mathematics also provided technical advice as Danielle's PhD co-advisor.  And Jennifer Van Boxel from Dicey Stories proof read an earlier version of this manuscript.

\appendix

\section{Benchmark Regression Posterior}\label{sec:post_det}

\begin{figure}[!h]
\centering
    \includegraphics[width=0.8\textwidth]{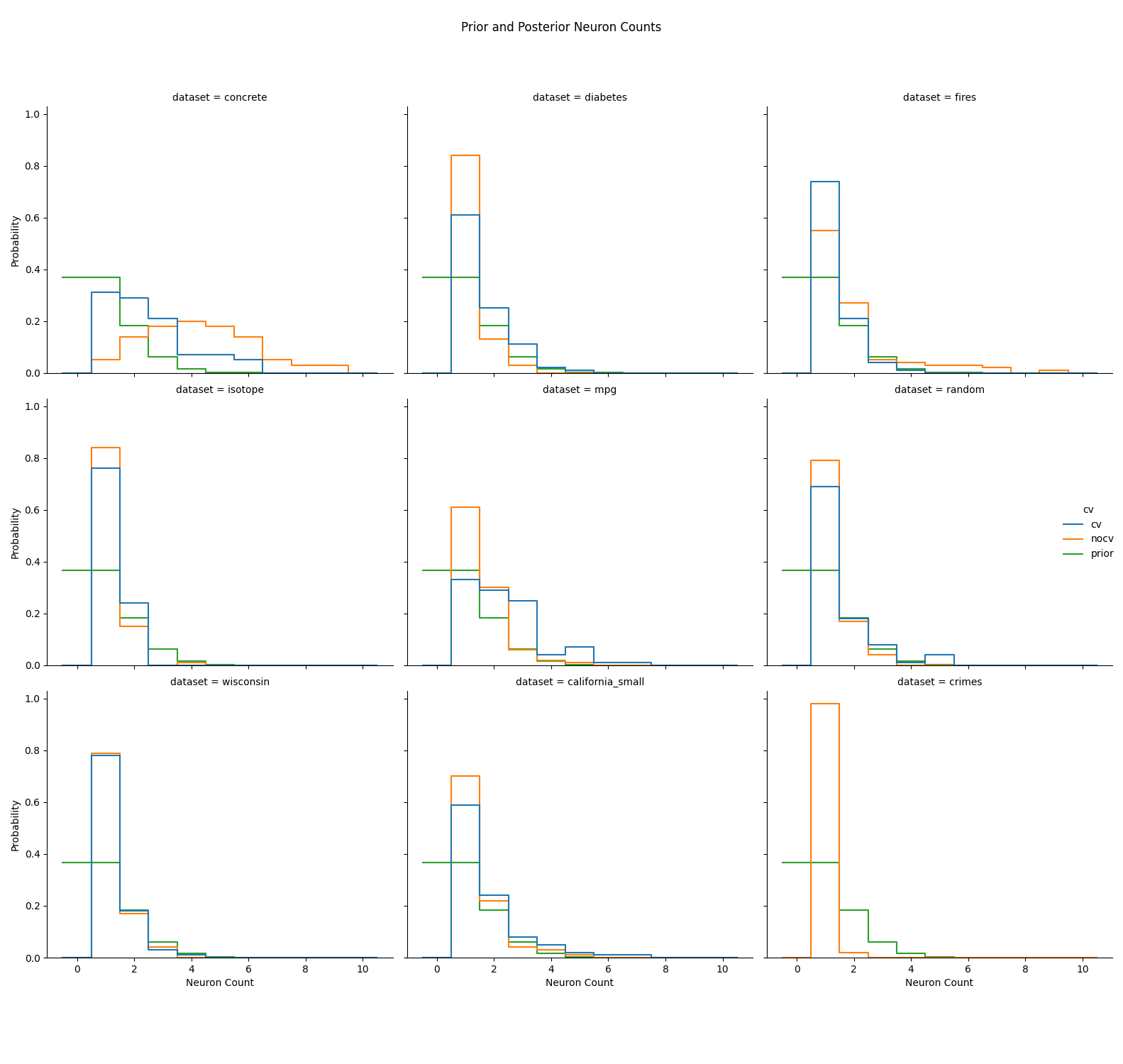}
    \caption{Posterior distribution of neuron counts varies across data sets}
    \label{fig:neuron_count}
\end{figure}

Here we briefly review the neuron count distributions within the BARN ensemble on different types of data.  \Cref{fig:neuron_count} shows the posterior distribution of neuron counts for both the tuned and untuned BARN methods.  In addition, we include on each graph the prior distribution (with the caveat that zero neurons was never actually possible).  For all but the concrete data set, the final distribution most often has a single neuron in each network (about 80\% of the time).  The concrete data is highly nonlinear, so it not surprising that BARN needs more neurons (2 to 4) per network to model this.

\section{Synthetic Regression Details}\label{sec:more_synth_reg}

\begin{figure*}[!h]
\centering
	\begin{subfigure}[t]{\textwidth}
		\centering
        \makebox[\textwidth][c]{\includegraphics[width=1.2\textwidth]{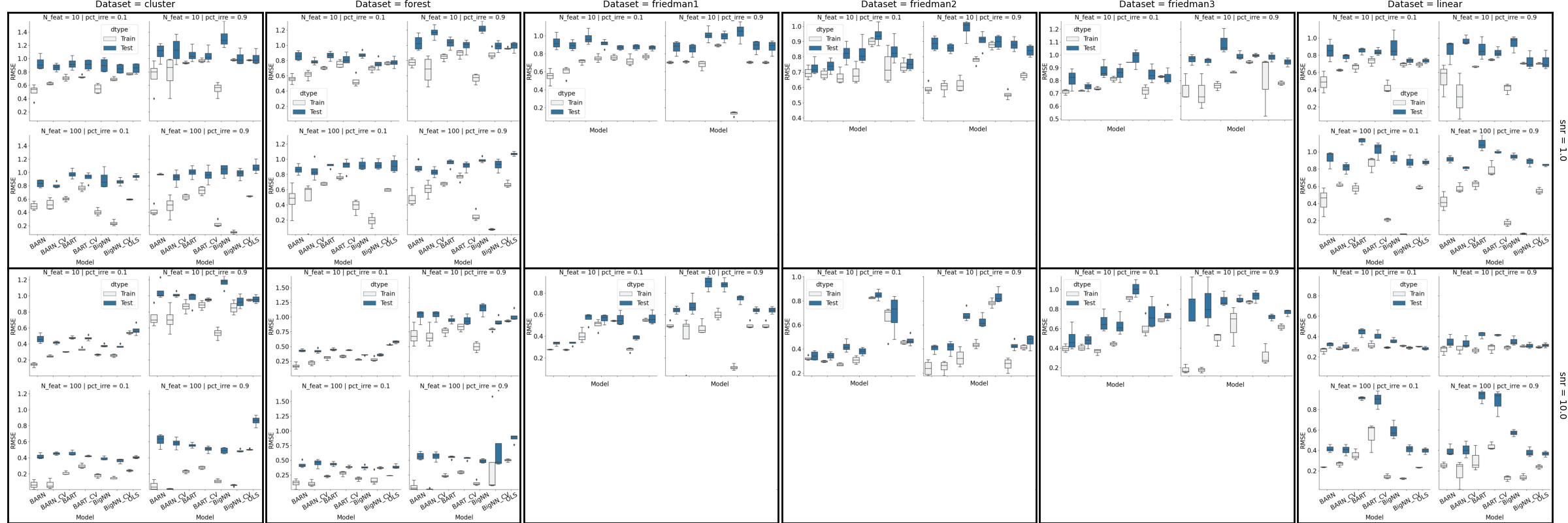}}
		\caption{Individual RMSE results, training and test}
		\label{fig:synth_box_matrix_results}
	\end{subfigure}%

	\begin{subfigure}[t]{\textwidth}
		\centering
        \makebox[\textwidth][c]{\includegraphics[width=1.2\textwidth]{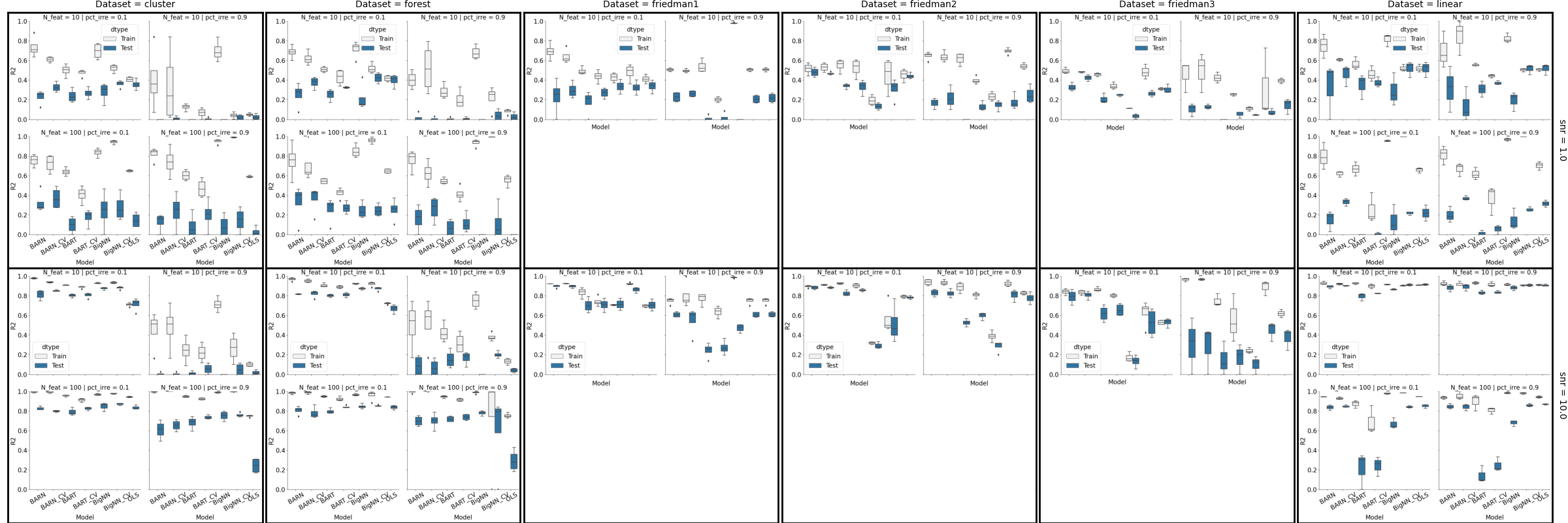}}
		\caption{Individual $R^2$ results, both training and test}
		\label{fig:synth_r2_matrix_results}
	\end{subfigure}
    \caption{Synthetic data results, grouped by data set and SNR level (heavy boxes), as well as number of features and irrelevant features (interior division)}
    \label{fig:synth_all_matrix_results}
\end{figure*}

We include some additional figures which go into more detail for individual data sets.  \Cref{fig:synth_all_matrix_results} shows the relative testing and training error results for each synthetic data set, organized by the factors that comprise them.

\clearpage

\bibliographystyle{elsarticle-harv} 
\bibliography{lib.bib}


%\begin{thebibliography}{00}


%\bibitem{}

%\end{thebibliography}
\end{document}